\documentclass[lettersize,journal]{IEEEtran}
\usepackage{amsmath,amsfonts}
\usepackage{algorithmic}
\usepackage{algorithm}
\usepackage{array}
\usepackage{multirow}
\usepackage[caption=false,font=normalsize,labelfont=sf,textfont=sf]{subfig}
\usepackage{textcomp}
\usepackage{stfloats}
\usepackage{url}
\usepackage{verbatim}
\usepackage{graphicx}
\usepackage{cite}
\usepackage{etoolbox}
\usepackage[monochrome]{color}
\begin{document}

\title{DiffCap: Diffusion-based Real-time \\ Human Motion Capture using Sparse IMUs \\ and a Monocular Camera}

\author{Shaohua Pan, Xinyu Yi, Yan Zhou, Weihua Jian, Yuan Zhang, Pengfei Wan and Feng Xu
\thanks{Shaohua Pan, Xinyu Yi, and Feng Xu are with Tsinghua University. Yan Zhou, Weihua Jian, Yuan Zhang, and Pengfei Wan are with Kuaishou Technology.}
}

\markboth{Journal of \LaTeX\ Class Files,~Vol.~14, No.~8, August~2021}%
{Shell \MakeLowercase{\textit{et al.}}: A Sample Article Using IEEEtran.cls for IEEE Journals}


\maketitle

\begin{abstract}
Combining sparse IMUs and a monocular camera is a new promising setting to perform real-time human motion capture.
This paper proposes a diffusion-based solution to learn human motion priors and fuse the two modalities of signals together seamlessly in a unified framework. 
By delicately considering the characteristics of the two signals, the sequential visual information is considered as a whole and transformed into a condition embedding, while the inertial measurement is concatenated with the noisy body pose frame by frame to construct a sequential input for the diffusion model.
Firstly, we observe that the visual information may be unavailable in some frames due to occlusions or subjects moving out of the camera view.
Thus incorporating the sequential visual features as a whole to get a single feature embedding is robust to the occasional degenerations of visual information in those frames.
On the other hand, the IMU measurements are robust to occlusions and always stable when signal transmission has no problem.
So incorporating them frame-wisely could better explore the temporal information for the system.
Experiments have demonstrated the effectiveness of the system design and its state-of-the-art performance in pose estimation compared with the previous works.
Our codes are available for research at https://shaohua-pan.github.io/diffcap-page.
\end{abstract}


\section{Introduction}
\IEEEPARstart{H}{uman} motion capture (mocap) has important applications in various fields such as sports, AR, VR, animation, and games.
Although commercial solutions (e.g., Vicon \cite{Vicon} and Xsens \cite{Xsens}) excel at accurately capturing human motion, their complexity and high cost present significant barriers for common users.
Nowadays, RGB cameras and inertial measurement units (IMUs) are commonly integrated into daily wearable devices and phones.
Leveraging them makes mocap much more convenient and may enable some new applications. 
In both academia and industry, more and more attention has been paid to pursue mocap with daily sensors.
\par
With these sensors, there are majorly three convenient ways to perform human motion capture, i.e., using monocular RGB camera (vision-based) \cite{spin, ROMP, li2021hybrik, PARE, VIBE, cliff}, using sparse IMUs (IMU-based) \cite{DIP, TransPose, PIP, TIP}, and using the combination of both (fusion-based) \cite{HybridCap, VIPpose, pan2023fusing}.
Vision-based methods have become increasingly efficient and effective, leveraged by the advances in deep learning techniques.
However, they fail when the visual signal is compromised, such as occlusion or people moving out of the camera view.
IMU-based methods do not suffer these limitations, but they are susceptible to drift issues due to error accumulation.
For the fusion-based methods, \cite{VIPpose, HybridCap} combine a monocular RGB camera with sparse IMUs to achieve more stable and accurate performance while maintaining convenience relatively.
However, \cite{VIPpose} faces efficiency challenges due to its reliance on sequence optimization, while \cite{HybridCap} does not account for the scenarios where performers may be occluded or move out of the camera view.
\cite{pan2023fusing} introduces a novel deep learning-based fusion method to overcome the aforementioned problems. 
However, it relies on the confidences of 2D keypoints (obtained from the monocular images by a 2D keypoints detector) to empirically switch between vision-combining-IMU model and pure IMU model, where fine-tuning the switching thresholds requires considerable effort, and its performance is significantly influenced by the accuracy of the 2D confidences.
There is still no effective solution to fuse the two kinds of information to achieve fast, robust, and accurate motion capture.
\par
The principal challenge in the fusion task arises from the time-varying confidences of the two types of signals. 
On one hand, the IMU signal accurately reflects body motion within a local time window. However, for long-term recordings, the signal unavoidably experiences drifting due to error accumulation.
On the other hand, the visual signal exhibits fluctuating confidence, and its degeneration is almost unpredictable as we cannot know when and which body parts suffer occlusions or user moving out of the camera view.
How to model the time-varying confidence of the two signals to maximize the information gain and suppress the errors is the key to achieve an ideal fusion-based mocap.  
\par
Inspired by the recent success of diffusion models for generating images \cite{ddpm} and motions \cite{mdm}, we propose to leverage diffusion models to perform this fusion task.
We believe that the learned prior knowledge of diffusion models contributes not only to the generation task but also to the reconstruction task.
Hence, we propose DiffCap, to leverage diffusion models in such a reconstruction scenario to perform robust human motion capture using sparse IMUs and a monocular camera.
By carefully designing the utilization of the two signals, we take into account the characteristics of the time-varying properties in solving for the human motion, and the solution space is constrained by the human motion prior knowledge learned by the diffusion model.
\par
DiffCap basically formulates the mocap task as a conditional generation task. 
It first transforms the sequence of 2D keypoints into a single-condition embedding which is utilized as the semantic information to guide the pose generation process of the diffusion model.
This single-condition embedding imparts low sensitivity to the availability of 2D keypoints in the sequence as the network can encode the visibility information in the embedding, enabling DiffCap to achieve a more robust mocap, especially for cases where the user is severely occluded or outside the camera view.
On the other hand, as the IMU measurements are always available in most cases, it will result in the loss of valuable temporal sequential information if considering them similarly with 2D keypoints as a single condition.
Therefore, the IMU measurements are retained at each frame, serving as input for the diffusion model.
Then the diffusion model will learn the correlation between the input IMUs and the output human motions. 
This maximizes the retention of input information and better handles situations when visual information is less reliable.
Furthermore, estimating human pose is a complex task due to the intricate nature of the rotation manifold (rotation of each joint in the human body skeleton), which is more complex than the estimation in Euclidean space (e.g., estimating body joint positions).
So we employ a two-stage solution with two diffusion models to reconstruct human motions: 1) Joint Diffusion Model is utilized to estimate human 3D joint positions, 2) Pose Diffusion Model is employed to estimate human pose from IMUs, 2D keypoints, and the 3D joint positions obtained from the Joint Diffusion Model.
\par
Experiments show that DiffCap achieves new state-of-the-art results on the AIST++ \cite{AIST++} (featuring challenging motion scenario), TotalCapture \cite{TotalCapture} (including scenarios with subjects out of the camera’s view), 3DPW \cite{VIPpose} (representing in-the-wild poses), and 3DPW-OCC \cite{ooh2020,VIPpose} (illustrating in-the-wild occluded poses) datasets.
We additionally show real-time live demos covering severe occlusion (e.g., umbrellas, and boards), out of camera view, challenging poses (e.g., dancing and boxing), and in-the-wild scenarios.
\begin{figure*}
\includegraphics[width=\linewidth]{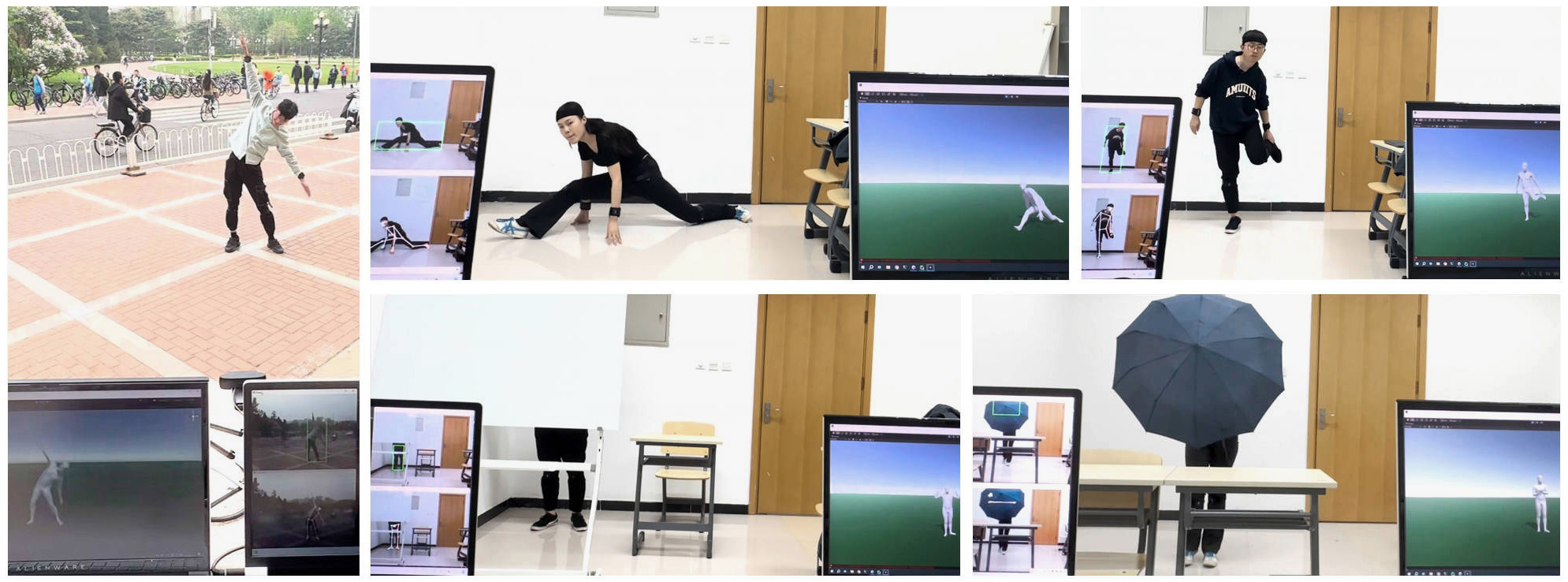}
  \centering
  \caption{DiffCap employs six inertial measurement units (IMUs) alongside a monocular camera to capture human motion robustly and accurately in real-time. The six IMUs are placed on the subject's left and right forearms, the left and right lower legs, the head, and the pelvis.
  This system effectively records human movement with high precision and reliability, even in challenging scenarios such as outdoor environments (left), complex motions (top right), and severe occlusions (bottom right).}
  \label{fig:teaser}
\end{figure*}
\begin{figure*}
  \includegraphics[width=\textwidth]{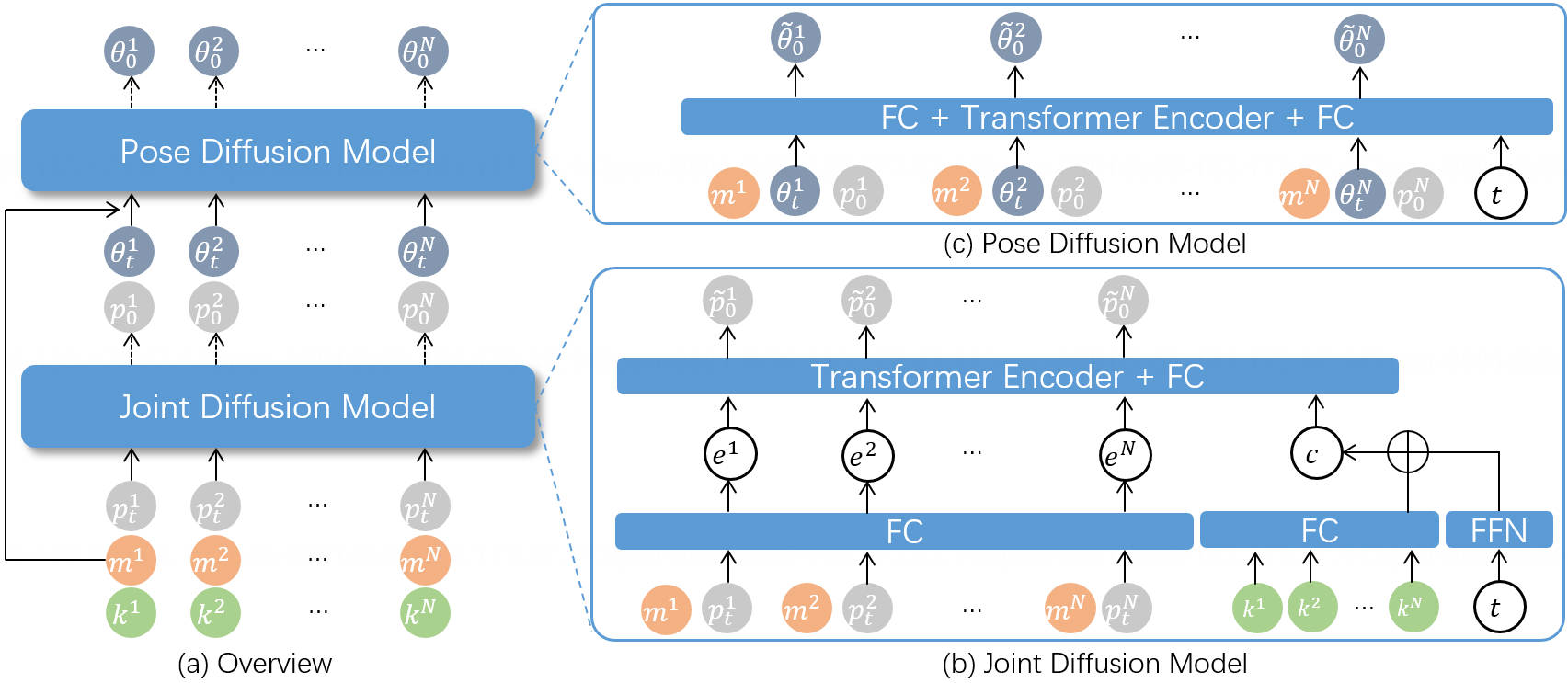}
  \caption{Our method. (a) An overview of our approach. (b) Joint Diffusion Model architecture. (c) Pose Diffusion Model architecture. The inputs consist of sequential 2D keypoints ($\boldsymbol{k}^{1:N}$) and the inertial measurements of 6 IMUs ($\boldsymbol{m}^{1:N}$). The outputs are the estimated sequential poses $\boldsymbol{\theta}_{0}^{1:N}$. ${t}$ represents the noising step. $\boldsymbol{p}_{{t}}^{1:N}$ and $\boldsymbol{\theta}_{{t}}^{1:N}$ denote the joint position and rotation sequence at step ${t}$, which are pure Gaussian noise when $t=T$. $\boldsymbol{e}^{1:N}$ denotes the sequential embedding inputted into the transformer encoder. $c$ is the condition embedding used to guide the diffusion denoising process. $\widetilde{\boldsymbol{p}}_{0}^{1:N}$ and $\widetilde{\boldsymbol{\theta}}_{0}^{1:N}$ are the results of each denoising step. After $T$ steps, $\boldsymbol{p}_{0}^{1:N}$ and $\boldsymbol{\theta}_{0}^{1:N}$ can be obtained from $\widetilde{\boldsymbol{p}}_{0}^{1:N}$ and $\widetilde{\boldsymbol{\theta}}_{0}^{1:N}$, respectively. We first utilize the Joint Diffusion Model to estimate joint positions and then use the Pose Diffusion Model to estimate human poses.}
  \label{fig:method}
\end{figure*}
\section{Related Work}
\subsection{Human Motion Capture}
We classify the works of human motion capture based on the input modality: vision-based, sensor-based, and vision-sensor fusion methods.
\par
{\it Vision-based Mocap:} In the domain of vision-based mocap, various approaches have shown impressive results by employing multi-view input \cite{harvesting17, CrossVF, LearnableTO, remelli2020lightweight, chun2023learnable, reddy2021tessetrack}.
However, such systems tend to be burdensome, and the recording space they require is limited.
Consequently, some approaches have turned to using a single monocular camera and applying optimization techniques \cite{Lassner, Bogo:ECCV:2016,xiang2019monocular,pavlakos2018learning} to refine the pose parameters represented by SMPL \cite{SMPL}.
Regression-based methods \cite{kolotouros2019convolutional, hmr, kanazawa2019learning, VIBE, zanfir2021neural, minimal, zhang2022mixste, PhysAware, ROMP, PARE, yuan2022glamr,BEV,huang2022occluded, Ma_2023_CVPR, li2021hybrik, li2023hybrik, li2023niki, motionbert2022, goel2023humans, shin2023wham} train deep neural networks to estimate the pose parameters directly from the input or some low-level features. 
Even though vision-based methods have shown remarkable performance, they are inherently susceptible to extreme lighting conditions and severe occlusions.
\par
{\it Sensor-based Mocap:} Unlike vision-based methods, sensor-based methods are not affected by challenging lighting conditions, occlusions, or limitations in the camera view.
Commercial inertial mocap solutions such as Xsens \cite{schepers2018xsens} have achieved high accuracy but require dense sensors (17 IMUs), making them inconvenient and expensive.  
Recently, mocap with sparse sensors has drawn much more attention.
\cite{SIP, DIP, TransPose, PIP, TIP, vanwouwe2023diffusion, yi2024pnp, armani2024ultra} use sparse inertial sensors to capture the human motion.
\cite{jiang2022avatarposer,winkler2022questsim,ye2022neural3points,aliakbarian2022flag,du2023agrol} leverage VR devices on the user's head and hands to estimate full-body motion.
Nevertheless, these methods encounter challenges such as pose ambiguity arising from the sparse sensor setting and the accumulation of sensor errors, leading to issues like jittery or drifting motion estimations.
\par
{\it Fusion-based Mocap:} Depending on the type of the visual input, the fusion-based methods can be categorized into three: fusing IMUs with multi-view video  \cite{TotalCapture,malleson2020real,RealTimeFM,gilbert2019fusing,FusingWI,moniruzzaman2021wearable,bao2022fusepose,huang2020deepfuse}, with RGBD video \cite{zheng2018hybridfusion}, and with monocular RGB video \cite{VIPpose,HybridCap,kaichi2020resolving,henschel2020accurate,henschel2019simultaneous, pan2023fusing,cha2021mobile,lee2024mocap}.
In the former two categories, \cite{vonmarcardponsmollPAMI16, RealTimeFM,pons2010multisensor,malleson2020real} use optimization techniques by minimizing energy functions that are correlated with both visual and inertial features. 
Other approaches \cite{TotalCapture,gilbert2019fusing,FusingWI,bao2022fusepose} involve the utilization of deep neural networks for pose estimation.  
In the case of methods combining monocular RGB with IMUs, the task becomes more challenging due to the inherent limitation of monocular RGB, which provides less information compared to RGBD or multi-view RGB.
\cite{henschel2020accurate,henschel2019simultaneous} only estimate the global translation and \cite{kaichi2020resolving} requires dense IMUs.
\textcolor{red}{\cite{lee2024mocap,cha2021mobile,EgoLocate} utilize a ‌body-mounted camera to capture the environment.
‌While this setup enhances localization accuracy, it is less effective for human motion prediction compared to configurations where the camera is focused on capturing the human subject.‌}
VIP \cite{VIPpose} utilizes an optimization-based approach to refine human pose and translation by incorporating 2D keypoints and inertial inputs.
However, VIP is designed as an offline technique and is not suitable for real-time applications.
HybridCap \cite{HybridCap} integrates information from 4 IMUs and 2D keypoint positions in the image domain to jointly estimate both human pose and translation.
However, HybridCap is designed to operate under the assumption that both types of inputs are available.
It may encounter challenges and even fail when the visual information is unavailable, as relying solely on 4 IMUs for motion capture can be difficult in such situations.
RobustCap \cite{pan2023fusing} employs a dual coordinate strategy to perform accurate and robust mocap from 6 IMUs and a monocular camera. 
However, RobustCap relies on the average confidence of 2D keypoints to determine the switch between the two coordinates.
Fine-tuning the parameters requires substantial effort, and the method is highly influenced by the accuracy of the 2D confidence, adding a level of sensitivity to the overall performance.
Different from the switching method of RobustCap, we employ a unified diffusion model to estimate the human pose under various conditions, which reduces complexity and leads to more accurate and robust motion capture.
\subsection{Diffusion Model}
Diffusion models \cite{ddpm, nichol2021improved} represent a category of likelihood-based generative models characterized by their focus on progressively learning to add noise to, and subsequently denoise, data.
Through the process of learning to reverse the diffusion steps that introduce noise in finite successive stages, diffusion models can generate samples that closely align with a given data distribution associated with the provided dataset.
Diffusion models have recently showcased remarkable results across a diverse range of generative tasks, extending from visual images \cite{diffbeatsgan, saharia2021image} to natural language \cite{austin2023structured, gong2022diffuseq, gong2023diffuseqv2}.
Leveraging their formidable capacity to model data priors, diffusion models can be employed to generate human motion based on textual \cite{mdm, mdmprior, ude, chen2023executing, yuan2023physdiff, jiang2023motiongpt} or musical conditions \cite{tseng2022edge, ude}.
Harnessing the capability of diffusion models to capture human priors, \cite{du2023agrol, vanwouwe2023diffusion} have successfully generated motion from sparse sensor inputs.
However, these methods are tend to motion generation tasks. 
\cite{gong2023diffpose, holmquist2023diffpose, zhang2024rohm} explore the generation of motion using either monocular images or 2D keypoints.
However, these methods encounter difficulties when dealing with severe occlusion or when individuals are out of camera.
Different from previous works, our work introduces the diffusion model specifically designed to achieve accurate and robust motion capture using sparse IMUs and a monocular camera.
\section{Method}
As illustrated in Fig.~\ref{fig:method}a, our system takes the sequential 2D keypoint detections $\boldsymbol{k}^{1:N}$ of a subject, and the synchronized orientation and acceleration measurements of 6 IMUs $\boldsymbol{m}^{1:N}$ at frames $1$ to $N$ as input.
The 6 IMUs are placed on the subject's left and right forearms, the left and right lower legs, the head, and the pelvis.
The output of our algorithm is the pose $\boldsymbol{\theta}_{0}^{1:N}$ of the subject, represented as SMPL \cite{SMPL} pose parameters.
Note that DiffCap also estimates the human global translation using the techniques of \cite{pan2023fusing}.
We adopt the mean human shape, following the same setting as \cite{pan2023fusing}.
A calibration step is performed to determine the extrinsics between the IMUs' coordinate system and the camera coordinate system (details are provided in the supplementary materials).
In the following, we first review the basic concepts of the diffusion model (Sec. ~\ref{subsec:diffusion}).
Then, we introduce how we apply it in our task of pose estimation (Sec. ~\ref{subsec:network}).
Finally, we discuss our inference strategy to achieve a smooth estimation of the long sequence without error accumulation (Sec. ~\ref{subsec:inference}).
\subsection{Diffusion Model}\label{subsec:diffusion}
Diffusion model comprises two fundamental processes: 1) a forward process that gradually adds Gaussian noise to data, and 2) a reverse process that learns to reconstruct clean data from noise \cite{ddpm}.
To be specific, the forward diffusion process is defined as: 
\begin{equation}\label{eq:eq1}
    q\left(\boldsymbol{x}_{t}^{1:N}\big|\boldsymbol{x}_{0}^{1:N}\right):=\mathcal{N}\left(\boldsymbol{x}_{t}^{1:N}; \sqrt {\overline{\alpha}}_{t}\boldsymbol{x}_{0}^{1:N},(1-\overline{\alpha}_{t})\boldsymbol{I}\right),
\end{equation}
where $\boldsymbol{x}$ is the data (could be the joint position $\boldsymbol{p}$ or rotation $\boldsymbol{\theta}$ in our task), $\overline{\alpha}_{t}:=\prod_{i=1}^t\alpha_{i}:=\prod_{i=1}^t(1-\beta_{i})$ and $\beta_{i}$ is the cosine noise variance schedule.
Utilizing the reparameterization trick, Eq. ~\ref{eq:eq1} can be reformulated as:
\begin{equation}\label{eq:eq2}     \boldsymbol{x}_{t}^{1:N}=\sqrt{\overline{\alpha}_{t}}\boldsymbol{x}_{0}^{1:N}+\sqrt{1-\overline{\alpha}_{t}}\epsilon,\quad \epsilon\sim\mathcal{N}(0,1).
\end{equation}
Here, $\boldsymbol{x}_t^{1:N}$ tends to be a random noise as $t=T$ and $T\rightarrow\infty$.
The inverse diffusion process is defined as:
\begin{equation}\label{eq:eq3}
    q\left(\boldsymbol{x}_{t-1}^{1:N}\big|\boldsymbol{x}_{t}^{1:N}, \boldsymbol{x}_{0}^{1:N}\right):=\mathcal{N}\left(\boldsymbol{x}_{t-1}^{1:N}; \boldsymbol{\mu}(\boldsymbol{x}_t^{1:N}, \boldsymbol{x}_0^{1:N},t),\sigma_t^2\boldsymbol{I}\right),
\end{equation}
where $\sigma_t^2$ follows a fixed schedule, and $\boldsymbol{\mu}$ can be reformulated as:
\begin{equation}
    \boldsymbol{\mu}(\boldsymbol{x}_t^{1:N}, \boldsymbol{x}_0^{1:N},t)=\frac{\sqrt {\alpha}_{t}(1-\overline{\alpha}_{t-1})}{1-\overline{\alpha}_t}\boldsymbol{x}_{t}^{1:N}+\frac{\sqrt{\overline{\alpha}_{t-1}}\beta_{t}}{1-\overline{\alpha}_t}\boldsymbol{x}_0^{1:N}.
\end{equation}
\par
During training, we sample at timestep $t$ and generate the corresponding noise data $\boldsymbol{x}_t^{1:N}$ by Eq. ~\ref{eq:eq2}.
A neural network, denoted as $f(\boldsymbol{x}_t^{1:N}, t, \boldsymbol{c})$ with $\boldsymbol{c}$ represents a condition, is then trained to execute the denoising task.
Following the approaches in \cite{du2023agrol,mdm,vanwouwe2023diffusion}, the denoising network $f$ predicts the clean data $\widetilde{\boldsymbol{x}}_0^{1:N}$ and we use the Mean Squared Error (MSE) loss during training.
\par
During the inference phase, the learned denoising network $f$ is first employed to process a random noise sample $\boldsymbol{x}_T^{1:N}$ and reconstruct the corresponding data sample $\widetilde{\boldsymbol{x}}_0^{1:N}$.
Then, given $\boldsymbol{x}_T^{1:N}$ and $\widetilde{\boldsymbol{x}}_0^{1:N}$, the resampling of $\boldsymbol{x}_{T-1}^{1:N}$ is performed using Eq. ~\ref{eq:eq3}.
By performing the denoising and resampling steps iteratively, we reconstruct the entire data sample $\boldsymbol{x}_0^{1:N}$ through a sequential denoising process $\boldsymbol{x}_T^{1:N} \rightarrow \boldsymbol{x}_{T-1}^{1:N} \rightarrow \cdot\cdot\cdot \rightarrow \boldsymbol{x}_0^{1:N}$.
\begin{figure}
\includegraphics[width=\linewidth]{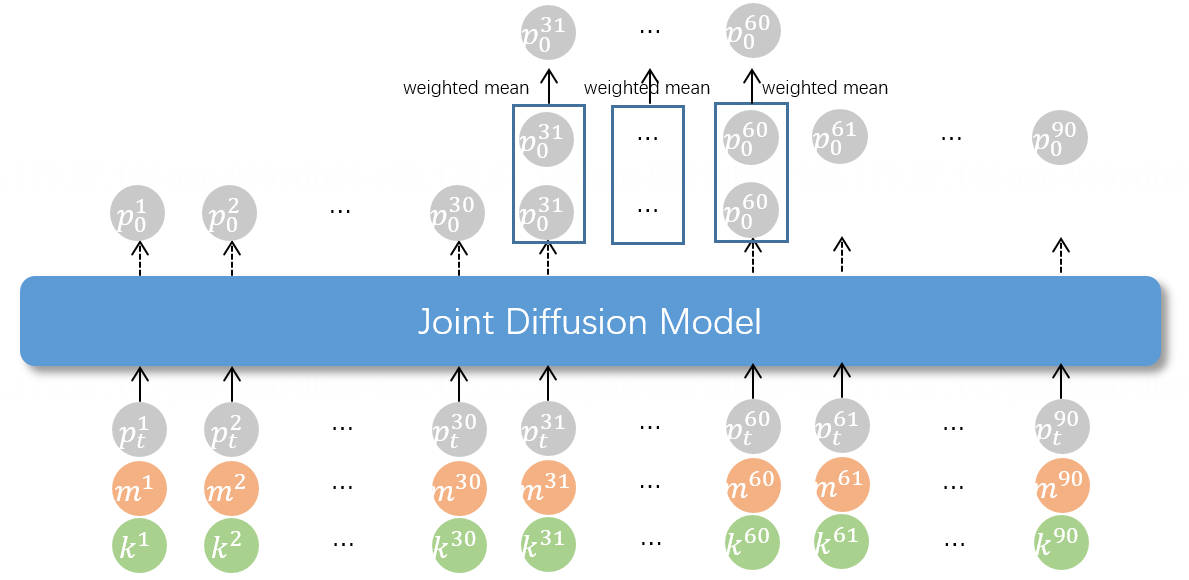}
  \centering
  \caption{Sliding window strategy during inference. First, We input the 60-frame sequence into the Joint Diffusion Model, obtaining $\boldsymbol{p}_0^{1:60}$. Next, we slide 30 frames and forward the Joint Diffusion Model to obtain $\boldsymbol{p}_0^{30:90}$. Finally, we perform a weighted average of the joint positions $\boldsymbol{p}_0^{31:60}$ to obtain the final results.}
  \label{fig:infer}
\end{figure}
\subsection{Diffusion-based Pose Estimation}\label{subsec:network}
As illustrated in Fig.~\ref{fig:method}, to mitigate the complexity of pose estimation, we decompose the pose estimation into two subtasks.
We first employ the Joint Diffusion Model to estimate the intermediate joint positions, then we utilize the Pose Diffusion Model to estimate the human pose.
In the following, we present the details of the Joint Diffusion Model and the Pose Diffusion Model.
\subsubsection{Joint Diffusion Model}\label{subsubsec:joint}
The Joint Diffusion Model $f_{p}$ is formulated as:
\begin{equation}
    \widetilde{\boldsymbol{p}}_0^{1:N} = f_{p}(\boldsymbol{p}_t^{1:N}, t, \boldsymbol{m}^{1:N}, \boldsymbol{k}^{1:N}),
\end{equation}
where $\boldsymbol{p}_t^{1:N}\in\mathbb{R}^{3JN}$ denotes the sequential joint positions at noise timestep $t$ ($J$ represents the number of joints and $N$ represents the number of frames), $\boldsymbol{m}^{1:N}\in\mathbb{R}^{72N}$ denotes the sequential acceleration and rotation measurements from the six IMUs, $\widetilde{\boldsymbol{p}}_0^{1:N}\in\mathbb{R}^{3JN}$ denotes the result of one denoising step, and $\boldsymbol{k}^{1:N}\in\mathbb{R}^{3KN}$ represents the preprocessed sequential 2D keypoints and their corresponding confidence ($K$ denotes the number of 2D keypoints).
\par
First, we discuss the 2D keypoints preprocessing process.
We use MediaPipe \cite{mediapipe} as our 2D keypoints detector.
It outputs the keypoints $\boldsymbol{p}_{2D}^{n}\in\mathbb{R}^{2K}$ and their corresponding confidence $\boldsymbol{\sigma^n}\in\mathbb{R}^{K}$ at frame $n$.
Next, We reproject the 2D keypoints onto the $Z = 1$ plane using the camera intrinsics to generalize to various camera settings.
Then we perform root normalization to represent the keypoints as root-relative 2D keypoints and the absolute position of the root in the camera coordinate system.
We concatenate the preprocessed 2D keypoints with their corresponding confidence to get $\boldsymbol{k}^{n}\in\mathbb{R}^{3K}$.
Finally, we concatenate $\boldsymbol{k}^{n}\in\mathbb{R}^{3K}$ of each frame to get the 2D keypoints input
$\boldsymbol{k}^{1:N}\in\mathbb{R}^{3KN}$.
\par
Then we discuss our network architecture.
The network architecture of the Joint Diffusion Model is depicted in Fig.~\ref{fig:method}b.
Following \cite{mdm}, we implement the Joint Diffusion Model using the transformer encoder architecture \cite{vaswani2023attention} to utilize temporal information.
$t$ and $\boldsymbol{k}^{1:N}$ are individually projected into the transformer latent dimension $D$ using a feed-forward network and a fully-connected layer, which are then summed to produce a condition embedding $\boldsymbol{c}\in\mathbb{R}^{D}$.
$\boldsymbol{c}$ extracts the global visual information (encoded in the 2D keypoints), which is then used in guiding the denoising process of the diffusion model.
Differently, the IMU measurements are concatenated with the noised joint positions of each frame, which are then projected into the transformer latent dimension individually, resulting in a sequential input embedding $\boldsymbol{e}^{1:N}\in\mathbb{R}^{DN}$.
Here, $\boldsymbol{e}^{1:N}$ contains the sequential IMU information with each frame's measurements explicitly encoded.
Then, $\boldsymbol{e}^{1:N}$ and $\boldsymbol{c}$ are concatenated and each component in the concatenated vector is summed with a standard positional embedding (the sum of position embedding is omitted in Fig. ~\ref{fig:method}) before feeding to the transformer encoder.
Excluding the last output token (corresponding to $\boldsymbol{c}$), the encoder output is linearly projected back to the original joint position dimensions, serving as the joint position prediction $\widetilde{\boldsymbol{p}}_0^{1:N}$.
\par
In this network design, the visual information is encoded globally while the inertial information is encoded locally with its original temporal ordering information.
The different designs are based on the key observation that the visual signals may be unavailable in the sequence due to the occlusion or subject moving out of the camera view, while the inertial signals are much more stable throughout the capturing process. 
For the visual information, if it is unavailable for some frames but we still encode it with the other information in these frames, it may easily pollute the output of these frames.
On the other hand, if we incorporate the visual information as a global guidance, the aforementioned problem is avoided and the visual information could still be utilized.
For the inertial information, since there is no availability problem, we could directly encode its local information in its original sequential order to maximize the utilization of the information.
The rationale behind this design choice is further demonstrated in Sec.~\ref{sec:inputeval}.
\subsubsection{Pose Diffusion Model}
The Pose Diffusion Model $f_{\theta}$ is formulated as:
\begin{equation}
    \widetilde{\boldsymbol{\theta}}_0^{1:N} = f_{\theta}(\boldsymbol{\theta}_t^{1:N}, t, \boldsymbol{m}^{1:N}, \boldsymbol{p}_0^{1:N}),
\end{equation}
where the $\boldsymbol{\theta}_t^{1:N}\in\mathbb{R}^{6JN}$ denotes the sequential joint rotations represented in 6D \cite{zhou2019continuity} at noise timestep $t$, ${\boldsymbol{p}}_0^{1:N}\in\mathbb{R}^{3JN}$ denotes the result of $T$ denoising step by the Joint Diffusion Model, and $\widetilde{\boldsymbol{\theta}}_0^{1:N}\in\mathbb{\theta}^{6JN}$ denotes the sequential joint rotations of one denoising step.
The network architecture of the Pose Diffusion Model is illustrated in Fig.~\ref{fig:method}c, which closely resembles the structure of the Joint Diffusion Model.
$\boldsymbol{m}^{1:N}$, $\boldsymbol{\theta}_t^{1:N}$, and $\boldsymbol{p}_0^{1:N}$ are linearly projected using the same layer.
After summing with positional embeddings, the combined information is fed into the transformer encoder.
The output is projected to the pose dimension, resulting in $\widetilde{\boldsymbol{\theta}}_0^{1:N}$.
\par 
In our pose estimation process, we intentionally exclude 2D keypoints from our inputs.
This design is based on the same observation mentioned in Sec. ~\ref{subsubsec:joint}, that 2D keypoints are less stable compared to IMUs.
Also, we have already incorporated the information from 2D keypoints in the Joint Diffusion Model.
Given our consideration that the IMU measurements remain stable during diverse motions, we introduce IMU as an additional input again to the Pose Diffusion Model.
This can be viewed as a skip connection technique, leveraging the stable information from IMUs to enhance the overall model performance.
\subsection{Inference}\label{subsec:inference}
We implement our diffusion model using a transformer encoder, and the transformer window size is fixed at 60.
During inference, we abstain from utilizing an auto-regressive strategy to prevent potential error accumulation.
We use the Joint Diffusion Model to illustrate our strategy, and the Pose Diffusion Model follows a similar approach to the Joint Diffusion Model.
As depicted in Fig. ~\ref{fig:infer}, we initially input the 60-frame sequence into the Joint Diffusion Model, obtaining $\boldsymbol{p}_0^{1:60}$.
Subsequently, we slide 30 frames and forward the Joint Diffusion Model to obtain $\boldsymbol{p}_0^{31:90}$.
Finally, we perform a weighted average of the joint positions $\boldsymbol{p}_0^{31:60}$ to obtain the final results.
\textcolor{red}{The weight of $\boldsymbol{p}_0^{31:60}$ in $\boldsymbol{p}_0^{1:60}$ linearly decreases from $1$ to $0$, while the weight of $\boldsymbol{p}_0^{31:60}$ in $\boldsymbol{p}_0^{31:90}$ linearly increases from $0$ to $1$.}
By employing this weighted mean, we achieve a smooth estimation of the long sequence without error accumulation.
\section{Experiments}
\begin{table*}[t]
\caption{Quantitative comparisons with state-of-the-art methods. We test our results on multiple datasets, including 3DPW test-split (representing in-the-wild poses), 3DPW-OCC (illustrating in-the-wild occluded poses), AIST++ test-split (featuring challenging motion scenarios), and TotalCapture (including scenarios with subjects out of the camera's view). For NIKI, VIP, and HybridCap, we report numerical results in their respective papers.RobustCap (biRNN) represents the application of bidirectional RNNs in RobustCap, allowing it to process the entire input sequence.}
\label{tab:allcmp}
\resizebox{\textwidth}{!}{
\begin{tabular}{cc|ccc|ccc|ccc|ccc}
\hline
\multicolumn{2}{c|}{}                                            & \multicolumn{3}{c|}{3DPW}                         & \multicolumn{3}{c|}{3DPW-OCC}                     & \multicolumn{3}{c|}{AIST++} & \multicolumn{3}{c}{TotalCapture} \\ \hline
\multicolumn{1}{c|}{Approach}                      & Method      & MPJPE & PA-MPJPE & PVE                            & MPJPE & PA-MPJPE & PVE                            & MPJPE  & PA-MPJPE  & PVE    & MPJPE    & PA-MPJPE    & PVE     \\ \hline
\multicolumn{1}{c|}{\multirow{3}{*}{Vision-based}} & ROMP        & 91.3  & 54.9     & 108.3                          & -     & -        & -                              & 90.3   & 60.0      & 128.1  & 145.4    & 64.6        & 184.5   \\
\multicolumn{1}{c|}{}                              & PARE        & 82.0  & 50.9     & 97.9                           & 90.5  & 56.6     & 107.9                          & 83.7   & 50.9      & 116.5  & 143.5    & 60.9        & 196.8   \\ 
\multicolumn{1}{c|}{}                              & NIKI        & 71.3  & 40.6     & 86.6                           & 85.5  & 53.5     & 107.6                          & -      & -         & -      & -        & -           & -     \\ \hline
\multicolumn{1}{c|}{\multirow{2}{*}{IMU-based}}    & PIP         & 78.0  & 49.8     & 100.0                          & 97.8  & 66.0     & 126.1                          & 87.1   & 62.0      & 116.5  & 49.1     & 34.6        & 66.0    \\
\multicolumn{1}{c|}{}                              & TIP         & 82.5  & 58.2     & 109.9                          & 100.5 & 68.7     & 131.6                          & 85.1   & 62.1      & 115.5  & 69.3     & 35.7        & 88.9    \\ \hline
\multicolumn{1}{c|}{\multirow{4}{*}{Fusion-based}}     & VIP         & -     & -        & -                              & -     & -        & -                              & -      & -         & -      & -        & 39.6        & -       \\
\multicolumn{1}{c|}{}                              & HybridCap   & 72.1  & -        & -                              & -     & -        & -                              & 33.3   & -         & -      & -        & -           & -       \\
\multicolumn{1}{c|}{}                              & RobustCap   & 55.0  & 38.9     & 71.8                           & 77.9  & 53.1     & 97.5                           & 33.1   & 24.0      & 43.2   & 48.7     & 33.5        & 63.4    \\
\multicolumn{1}{c|}{}                       & RobustCap (biRNN)  & 53.6  & 38.6     & 69.7                           & 76.8  & 52.5     & 96.2                           & 32.0   & 22.8      & 41.5   & 48.1     & 32.7        & 62.1     \\
\multicolumn{1}{c|}{} & \textbf{DiffCap} & \textbf{46.9}  & \textbf{33.5} & \textbf{65.9} & \textbf{60.5} & \textbf{43.1} & \textbf{85.6} & \textbf{31.0} & \textbf{21.2} & \textbf{40.5} & \textbf{46.2} & \textbf{29.9} & \textbf{60.9}    \\ \cline{1-14} 
\end{tabular}}
\end{table*}
In this section, we first elaborate on our implementation, and discuss the datasets and metrics (Sec.~\ref{sec:implementation}).
Then we compare our method with previous vision-based, IMU-based, and fusion-based approaches (Sec.~\ref{sec:comparisons}).
Next, we conduct ablation studies on our method (Sec.~\ref{sec:evaluations}).
Finally, we discuss the implementation and results of our live demo (Sec.~\ref{sec:livedemo})..
\subsection{Implementation Details}\label{sec:implementation}
All training and evaluation processes run on a computer with an Intel(R) Core(TM) i7-8700 CPU and an NVIDIA GTX2080Ti graphics card.
We build the transformer encoder network with 8 layers and 4 heads.
All latent features in the transformer encoder share a uniform length of 512, with a window size of $N=60$ and a dropout rate set at 0.1.
The network is trained with batch size 64 and AdamW optimizer \cite{adamw}.
The learning rate is set to $10^{-4}$.
The diffusion timestep is set to $T=1000$ during training and is set to 5 during sampling by DDIM \cite{ddim}.
Following RobustCap, we use MediaPipe \cite{mediapipe} as our 2D keypoints detector.
During training, the Pose Diffusion Model utilizes the 3D joint positions results estimated by the Joint Diffusion Model.
\paragraph{Datasets}
Our training runs on the AMASS dataset \cite{AMASS} and the AIST++ dataset \cite{AIST++}.
Following \cite{pan2023fusing}, for the AIST++ dataset we use detected 2D keypoints and synthesized IMUs, and for AMASS, we use synthesized 2D keypoints and IMUs.
We perform evaluations on AIST++ test split, TotalCapture \cite{TotalCapture}, 3DPW test split \cite{VIPpose}, and 3DPW-OCC \cite{VIPpose, ooh2020}.
We utilize synthesized IMUs for the evaluation of AIST++, 3DPW, and 3DPW-OCC datasets.
More details of the processing of the datasets are provided in the supplementary materials.
\paragraph{Metrics}
Following \cite{pan2023fusing}, we use similar metrics to evaluate our method.
1) MPJPE denotes the mean per joint position error in mm.
2) PA-MPJPE denotes the Procrustes-aligned mean per joint position error in mm.
3) PVE denotes the per-vertex error in mm using the SMPL mesh.
For all the metrics, the lower, the better.
\begin{figure*}
\includegraphics[width=\linewidth]{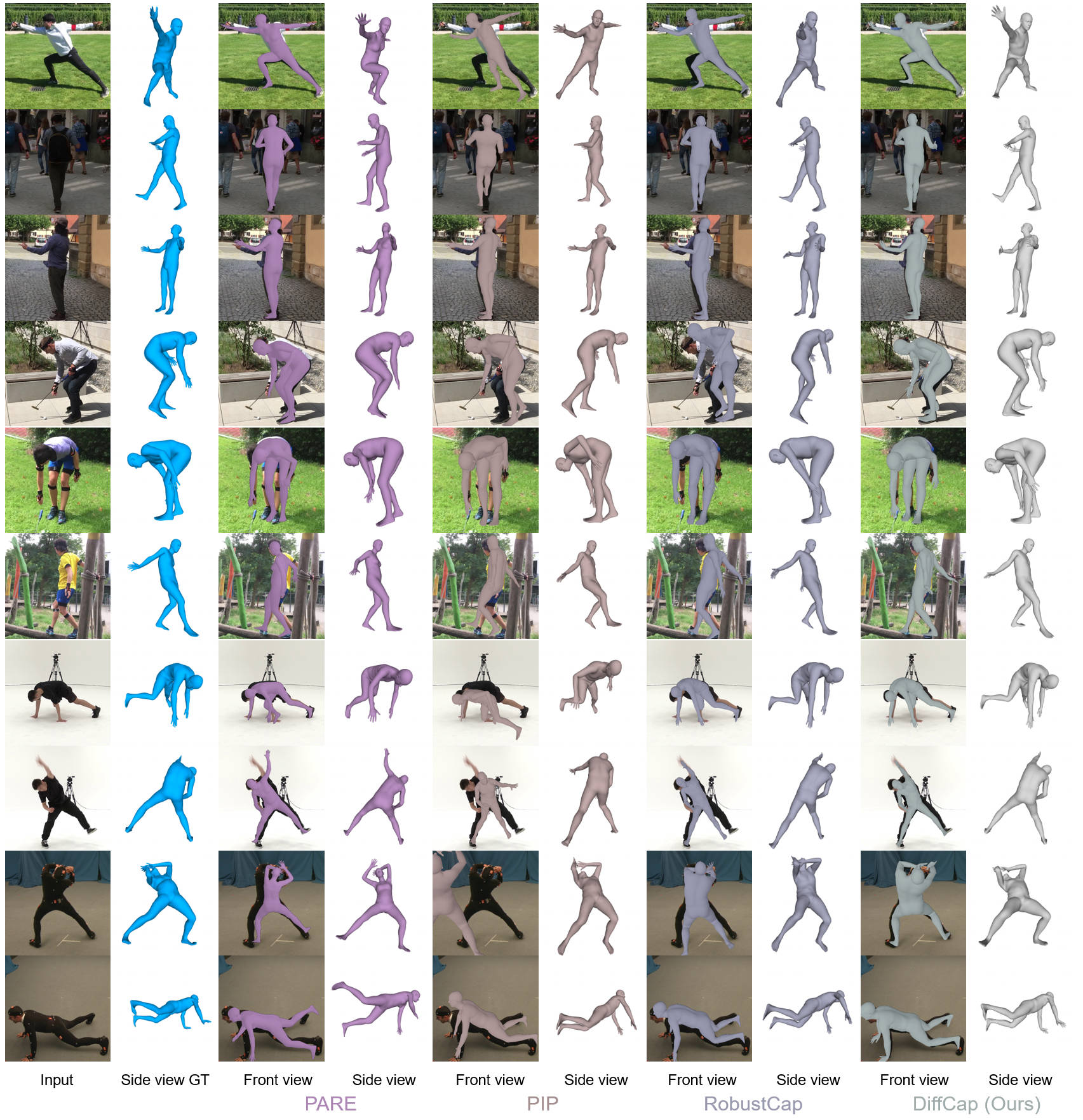}
  \centering
  \caption{Qualitative results on 3DPW (rows 1-3), 3DPW-OCC (rows 4-6), AIST++ (rows 7-8), and TotalCapture (rows 9-10) datasets. From left to right: input images, ground truth poses, results of a vision-based method (PARE), results of an IMU-based method (PIP), results of a fusion-based method (RobustCap), and Diffcap results.}
  \label{fig:comparison}
\end{figure*}
\begin{figure}
\includegraphics[width=\linewidth]{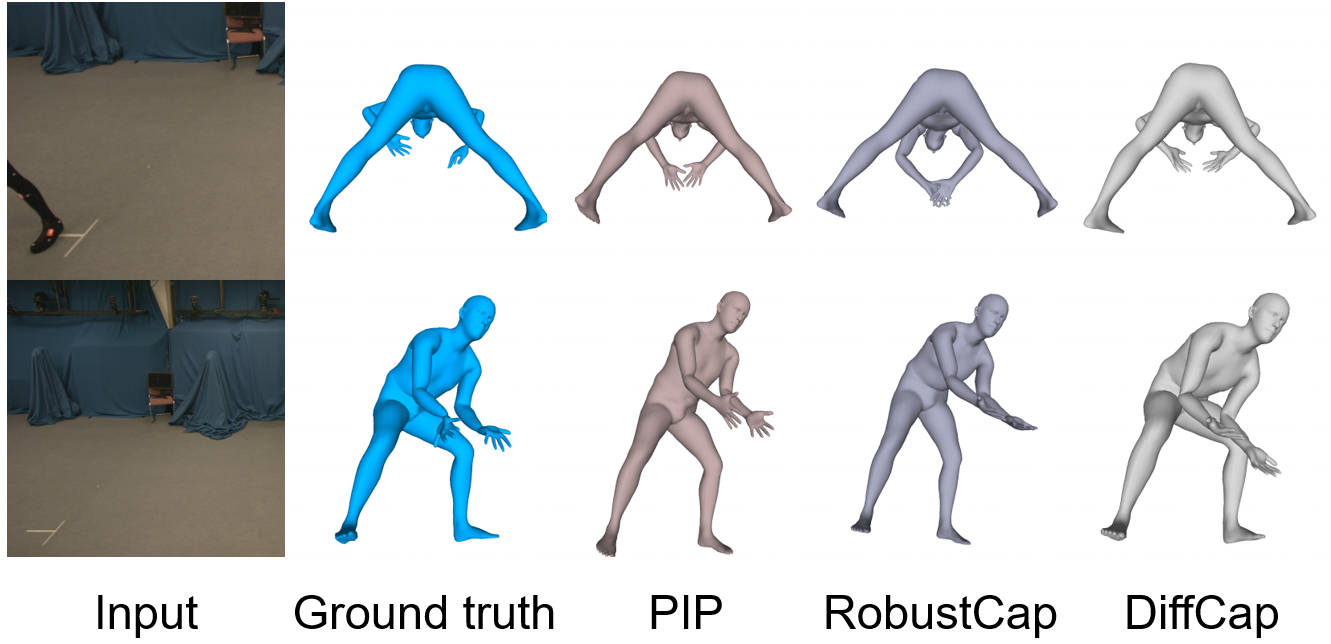}
  \centering
  \caption{
   Qualitative results on the TotalCapture dataset where the subjects are out of the camera view. Vision-based methods cannot estimate the human pose under this condition. IMU-based method PIP and fusion-based method RobustCap incorrectly estimate the arm poses in row 1. Additionally, their results in row 2 do not exhibit the expected stooped pose compared to the ground truth.}
  \label{fig:comparison2}
\end{figure}
\subsection{Comparisons}\label{sec:comparisons}
We compared our method with the state-of-the-art vision-based methods ROMP \cite{ROMP}, PARE \cite{PARE}, and NIKI \cite{li2023niki}, the IMU-based methods TIP \cite{TIP} and PIP \cite{PIP}, as well as the fusion-based methods HybridCap \cite{HybridCap}, VIP \cite{VIPpose}, and RobustCap \cite{pan2023fusing}. We also re-train RobustCap using bidirectional RNNs, enabling it to process the entire input sequence.
\subsubsection{Quantitative Comparisons}
The quantitative results are shown in Tab. ~\ref{tab:allcmp}.
Our method demonstrates significant improvements in pose accuracy compared to all the previous works.
Despite not being trained on the in-the-wild datasets, our method achieves top performance in terms of results on 3DPW and 3DPW-OCC which contain in-the-wild data and occlusion scenarios, demonstrating the generalization capability of our approach.
In demonstrating our method's adeptness at capturing challenging motions, we utilized the AIST++ dataset, featuring subjects performing a broad spectrum of complex dance movements within the camera's field of view.
Among the evaluated techniques, DiffCap achieves the greatest reduction in reconstruction errors, outperforming both vision-based and IMU-based methods, and also surpassing fusion-based approaches HybirdCap and RobustCap.
The TotalCapture dataset is utilized to depict situations in which subjects are outside the camera's field of view.
Vision-based methods fail in such instances.
Our novel diffusion-based architecture demonstrates superior performance in minimizing pose errors compared to existing IMU-based and fusion-based techniques.
Note that though DiffCap does not estimate the human shape, it still accomplishes the lowest PVE, demonstrating more accurate pose estimation.
DiffCap processes only 30 future frames as input, yet it outperforms the state-of-the-art RobustCap (biRNN), which uses the entire sequence.
\subsubsection{Qualitative Comparisons}
In Fig.~\ref{fig:comparison}, we show the qualitative results of DiffCap compared to the vision-based method PARE, the IMU-based method PIP, and the fusion-based method RobustCap, which are all state-of-the-art techniques of their categories.
The vision-based method PARE occasionally provides inaccurate estimations for end joints due to limited pixel information, thereby impeding the guidance of the vision-based approach, as depicted in the 4th, 5th, 6th, and 7th rows of Fig.~\ref{fig:comparison}.
Moreover, it yields incorrect estimations for blurred image inputs, as illustrated in the 8th row of Fig.~\ref{fig:comparison}.
On the other hand, the IMU-based method PIP experiences drift, as indicated by the worse overlay as shown in the 9th row of Fig.~\ref{fig:comparison}.
Furthermore, in the absence of image information, PIP encounters challenges in accurately estimating ambiguous poses, as evident in the 1st, 7th, and 8th rows of Fig.~\ref{fig:comparison}.
The fusion-based method RobustCap achieves superior results compared to PARE and PIP.
However, its performance is slightly compromised in certain cases, as suggested by the slightly inferior overlay.
RobustCap relies on a straightforward fusion strategy of linear interpolation between visual and inertial results.
And the interpolation is determined by the predicted confidence of the 2D keypoints, which may not be stable and accurate.
In contrast, our method encodes the sequence of 2D keypoints and their confidence into a latent space for signal fusion, which makes our method more robust to the noise of visual signals than RobustCap. 
Besides, our approach leverages vision and IMU inputs more effectively than previous works through the diffusion model.
Consequently, we can achieve superior pose estimation in challenging scenarios involving complex motions, occlusion, and in-the-wild conditions.
In Fig.~\ref{fig:comparison2}, we show the qualitative comparison results when the subject is out of the camera view.
Vision-based methods fail as the visual signal is not available.
With the learned human motion prior from of diffusion model, our pose estimation results surpass those of the IMU-based method PIP and the fusion-based method RobustCap, even in the absence of visual information.
\begin{table}[t]
\caption{Quantitative evaluation of the two-stage solution. One-stage refers to use the diffusion model to estimate human pose directly, while two-stage involves estimating 3D joint positions before poses.}
\label{tab:stage}
\resizebox{\columnwidth}{!}{
\begin{tabular}{c|ccc|ccc}
\hline
\multirow{2}{*}{Method} & \multicolumn{3}{c|}{AIST++}                       & \multicolumn{3}{c}{3DPW-OCC}                             \\ \cline{2-7} 
    & \multicolumn{1}{c}{MPJPE} & \multicolumn{1}{c}{PA-MPJPE} & \multicolumn{1}{c|}{PVE}     & \multicolumn{1}{c}{MPJPE} & \multicolumn{1}{c}{PA-MPJPE} & PVE \\ \hline
one-stage     & \multicolumn{1}{c}{35.4}  & \multicolumn{1}{c}{24.3}     & 45.7   & \multicolumn{1}{|c}{71.3}  & \multicolumn{1}{c}{51.3}     & 94.1            \\ 
two-stage  & \multicolumn{1}{c}{\textbf{31.0}} & \multicolumn{1}{c}{\textbf{21.2}} & \textbf{40.5} & \multicolumn{1}{|c}{\textbf{60.5}} & \multicolumn{1}{c}{\textbf{43.1}} & \textbf{85.6}   \\ \hline
\end{tabular}}
\end{table}
\begin{table}[t]
\caption{Quantitative evaluation of the strategies for inputting IMU and 2D keypoints to the diffusion model.}
\label{tab:input}
\resizebox{\columnwidth}{!}{
\begin{tabular}{c|ccc|ccc}
\hline
\multirow{2}{*}{Method} & \multicolumn{3}{c|}{AIST++}                       & \multicolumn{3}{c}{3DPW-OCC}                             \\ \cline{2-7} 
                        & \multicolumn{1}{c}{MPJPE} & \multicolumn{1}{c}{PA-MPJPE} & \multicolumn{1}{c}{PVE}      & \multicolumn{1}{|c}{MPJPE} & \multicolumn{1}{c}{PA-MPJPE} & PVE \\ \hline
both as condition     & \multicolumn{1}{c}{52.3}  & \multicolumn{1}{c}{35.2}     & \multicolumn{1}{c}{67.9}     & \multicolumn{1}{|c}{67.8}  & \multicolumn{1}{c}{47.2}     & 90.6            \\ 
both as sequential input         & \multicolumn{1}{c}{32.0}  & \multicolumn{1}{c}{22.3}     & \multicolumn{1}{c}{42.0}     & \multicolumn{1}{|c}{71.1}  & \multicolumn{1}{c}{49.3}     & 92.1   \\
DiffCap     & \multicolumn{1}{c}{\textbf{31.0}} & \multicolumn{1}{c}{\textbf{21.2}} & \textbf{40.5} & \multicolumn{1}{|c}{\textbf{60.5}} & \multicolumn{1}{c}{\textbf{43.1}} & \textbf{85.6}   \\ \hline
\end{tabular}}
\end{table}
\begin{figure}
\includegraphics[width=\linewidth]{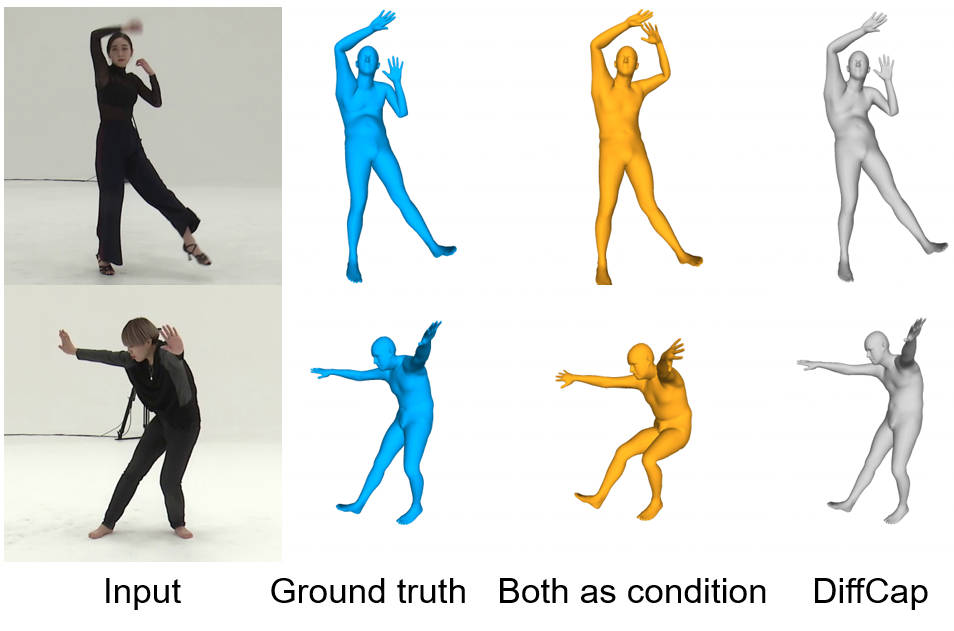}
  \centering
  \caption{Qualitative evaluation of the strategies for inputting IMU and 2D keypoints to the diffusion model. Both as condition denotes transforming both 2D keypoints and IMU into a condition embedding to guide the diffusion denoising process.}
  \label{fig:eval2}
\end{figure}
\begin{figure}
\includegraphics[width=\linewidth]{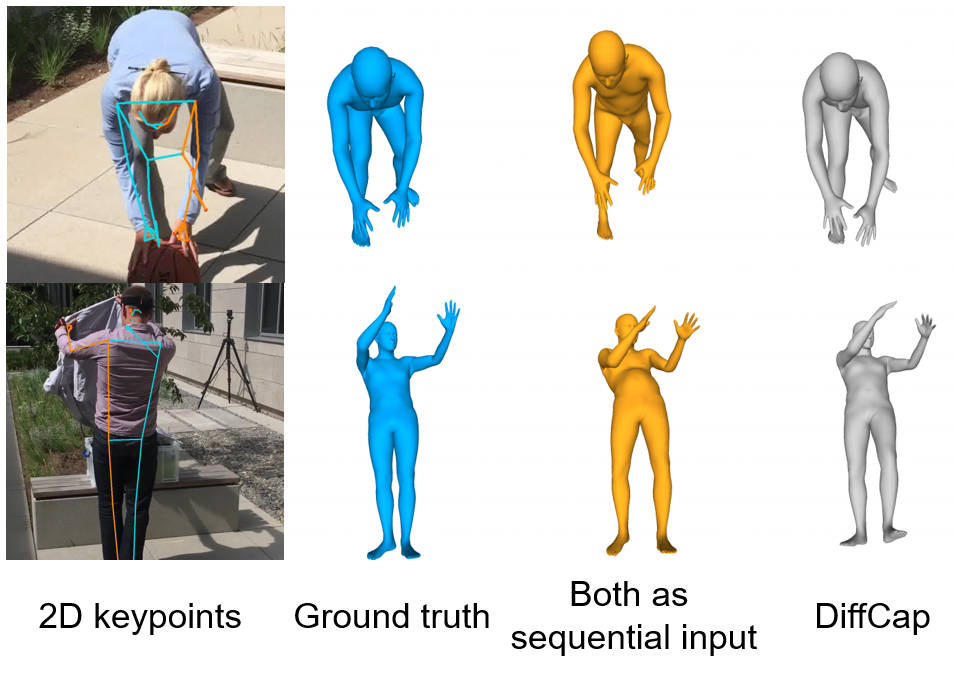}
  \centering
  \caption{Qualitative evaluation of the strategies for inputting IMU and 2D keypoints to the diffusion model. Both as sequential input denotes concatenating the 2D keypoints, IMU measurement with the noised 3D joint position at each frame as sequential input embeddings.}
  \label{fig:eval1}
\end{figure}
\subsection{Evaluations}\label{sec:evaluations}
In this section, we first evaluate the two-stage solution, which involves estimating joint positions before estimating poses.
Next, we explore different strategies for feeding inputs into the diffusion model.
We then examine the impact of the number of sampling steps during inference.
Following that, we assess the robustness of our model against noise in the 2D keypoints.
We also evaluate the effectiveness of the diffusion model.
Finally, we investigate the effect of varying the sliding window size in the sliding window strategy.
\subsubsection{Two-stage Solution}
For this experiment, we employ the same configuration to train a diffusion model for direct pose estimation.
As depicted in Tab. ~\ref{tab:stage}, the strategy of estimating joint positions first, rather than directly estimating human pose, is proved to be effective in reducing complexity, resulting in superior results.
Notice that the two-stage pose estimation still satisfies real-time applications.
\subsubsection{Sequential Input or Condition}\label{sec:inputeval}
We study various strategies for inputting IMUs and 2D keypoints to the diffusion model.
As illustrated in Sec. ~\ref{subsubsec:joint}, we could use different embedding methods for visual and inertial signals.
For 2D keypoints, we transform all frames within the transformer's time window into a condition embedding, incorporating it as global semantic information.
Additionally, we concatenate IMU measurements with the noised 3D joint positions for each frame to construct sequential input embedding during the denoising process.
We compare DiffCap with two alternatives:
1) \textit{Both as condition} denotes transforming all frames of 2D keypoints and IMU measurements within the transformer's time window into a condition embedding.
2) \textit{Both as sequential input} means concatenating the 2D keypoints and IMU measurements with the noised 3D joint positions frame-by-frame as sequential input embeddings during the diffusion denoising process.
\par
As shown in Tab.~\ref{tab:input}, the results indicate that compressing sequential IMU measurements as a condition (i.e., both as condition) leads to information loss, resulting in significantly worse outcomes compared to DiffCap in both datasets.
We additionally show two sequences from the AIST++ dataset, and the corresponding results are presented in Fig.~\ref{fig:eval2}.
The results of using both signals as conditions are inaccurate in some challenging poses.
On the other hand, when utilizing both 2D keypoints and IMU measurements as sequential input, the results in AIST++ (where performers are all within the camera's view) are comparable to DiffCap, which is indicated in Tab. ~\ref{tab:input}.
However, the reconstruction error experiences a significant increase in the 3DPW-OCC dataset (where performers are subject to severe occlusion) due to the errors in 2D keypoints, which misleads the predictions of the diffusion model.
We show two sequences from the 3DPW-OCC dataset and present the results in Fig.~\ref{fig:eval1}.
The results of using both signals as sequential input are inaccurate because of the wrong 2D keypoint positions estimated by MediaPipe (the hip in the first row and the right elbow in the second row).
In contrast, DiffCap is not sensitive to the noise of the 2D keypoints and achieves stable and accurate results.
The strategy employed by DiffCap maximizes the retention of input information and proves more adept at handling situations where 2D information is less reliable.
\begin{table}[t]
\caption{Quantitative evaluation of the number of diffusion sampling steps during inference.}
\label{tab:steps}
\resizebox{\columnwidth}{!}{
\begin{tabular}{c|ccc|ccc}
\hline
\multirow{2}{*}{\# steps} & \multicolumn{3}{c|}{AIST++}                       & \multicolumn{3}{c}{3DPW-OCC}                             \\ \cline{2-7} 
                        & \multicolumn{1}{c}{MPJPE} & \multicolumn{1}{c}{PA-MPJPE} & \multicolumn{1}{c}{PVE}      & \multicolumn{1}{|c}{MPJPE} & \multicolumn{1}{c}{PA-MPJPE} & PVE \\ \hline
1     & \multicolumn{1}{c}{32.7}  & \multicolumn{1}{c}{22.0}     & \multicolumn{1}{c}{42.5}     & \multicolumn{1}{|c}{60.8}  & \multicolumn{1}{c}{\textbf{43.1}}     & 85.8            \\ 
5     & \multicolumn{1}{c}{\textbf{31.0}}  & \multicolumn{1}{c}{\textbf{21.2}}     & \multicolumn{1}{c}{\textbf{40.5}}     & \multicolumn{1}{|c}{\textbf{60.5}} & \multicolumn{1}{c}{\textbf{43.1}} & \textbf{85.6}   \\
10    & \multicolumn{1}{c}{\textbf{31.0}}  & \multicolumn{1}{c}{\textbf{21.2}}     & \multicolumn{1}{c}{40.6}     & \multicolumn{1}{|c}{60.6}  & \multicolumn{1}{c}{43.3}     & 85.7   \\ \hline
\end{tabular}}
\end{table}

\begin{table}[t]
\caption{Quantitative evaluation on the robustness of DiffCap. We introduce Gaussian noise ($\sigma$) to the 2D keypoints (on $Z=1$ plane) of AIST++ while maintaining the confidence. Compared to RobustCap, DiffCap is less sensitive to the noise in 2D keypoints.}
\label{tab:robustness}
\resizebox{\columnwidth}{!}{
\begin{tabular}{c|ccc|ccc}
\hline
\multirow{2}{*}{\# sigma} & \multicolumn{3}{c|}{RobustCap}                       & \multicolumn{3}{c}{DiffCap}                             \\ \cline{2-7} 
                        & \multicolumn{1}{c}{MPJPE} & \multicolumn{1}{c}{PA-MPJPE} & \multicolumn{1}{c}{PVE}      & \multicolumn{1}{|c}{MPJPE} & \multicolumn{1}{c}{PA-MPJPE} & PVE \\ \hline
0     & \multicolumn{1}{c}{33.1}  & \multicolumn{1}{c}{24.0}     & \multicolumn{1}{c}{43.2}     & \multicolumn{1}{|c}{\textbf{31.0}}  & \multicolumn{1}{c}{\textbf{21.2}}     & \textbf{40.5}            \\ 
\textcolor{red}{0.01}  & \multicolumn{1}{c}{35.6}  & \multicolumn{1}{c}{25.9}     & \multicolumn{1}{c}{46.2}     & \multicolumn{1}{|c}{31.2} & \multicolumn{1}{c}{21.4}     & 40.8   \\
\textcolor{red}{0.05}  & \multicolumn{1}{c}{41.0}  & \multicolumn{1}{c}{29.9}     & \multicolumn{1}{c}{53.5}     & \multicolumn{1}{|c}{32.0} & \multicolumn{1}{c}{22.1}     & 41.8   \\
0.1   & \multicolumn{1}{c}{42.3}  & \multicolumn{1}{c}{31.0}     & \multicolumn{1}{c}{55.4}     & \multicolumn{1}{|c}{32.7} & \multicolumn{1}{c}{22.0}     & 42.5   \\
\textcolor{red}{0.2}   & \multicolumn{1}{c}{43.0}  & \multicolumn{1}{c}{31.5}     & \multicolumn{1}{c}{56.3}     & \multicolumn{1}{|c}{33.2} & \multicolumn{1}{c}{22.4}     & 43.1   \\
\hline
\end{tabular}}
\end{table}
\begin{table}[t]
\caption{Quantitative evaluation of the diffusion model. W/o diffusion model denotes the substitution of the diffusion model with transformer encoders.}
\label{tab:diffusion}
\resizebox{\columnwidth}{!}{
\begin{tabular}{c|ccc|ccc}
\hline
\multirow{2}{*}{Method} & \multicolumn{3}{c|}{AIST++}                       & \multicolumn{3}{c}{3DPW-OCC}                             \\ \cline{2-7} 
                        & \multicolumn{1}{c}{MPJPE} & \multicolumn{1}{c}{PA-MPJPE} & \multicolumn{1}{c}{PVE}      & \multicolumn{1}{|c}{MPJPE} & \multicolumn{1}{c}{PA-MPJPE} & PVE \\ \hline
w/o diffusion model     & \multicolumn{1}{c}{35.2}  & \multicolumn{1}{c}{23.7}     & \multicolumn{1}{c}{45.4}     & \multicolumn{1}{|c}{70.7}  & \multicolumn{1}{c}{49.4}     & 95.5            \\ 
\textcolor{red}{w/o pose diffusion model}  & \multicolumn{1}{c}{33.0}  & \multicolumn{1}{c}{22.3}     & \multicolumn{1}{c}{43.0}     & \multicolumn{1}{|c}{65.3}  & \multicolumn{1}{c}{47.2}     & 91.0            \\ 
DiffCap     & \multicolumn{1}{c}{\textbf{31.0}} & \multicolumn{1}{c}{\textbf{21.2}} & \textbf{40.5} & \multicolumn{1}{|c}{\textbf{60.5}} & \multicolumn{1}{c}{\textbf{43.1}} & \textbf{85.6}   \\ \hline
\end{tabular}}
\end{table}
\begin{table}[t]
\caption{Quantitative evaluation of the number of sliding step during inference.}
\label{tab:window-size}
\resizebox{\columnwidth}{!}{
\begin{tabular}{c|ccc|ccc}
\hline
\multirow{2}{*}{\# sliding step} & \multicolumn{3}{c|}{3DPW}                       & \multicolumn{3}{c}{3DPW-OCC}                             \\ \cline{2-7} 
                        & \multicolumn{1}{c}{MPJPE} & \multicolumn{1}{c}{PA-MPJPE} & \multicolumn{1}{c}{PVE}      & \multicolumn{1}{|c}{MPJPE} & \multicolumn{1}{c}{PA-MPJPE} & PVE \\ \hline
10     & \multicolumn{1}{c}{48.1}  & \multicolumn{1}{c}{34.5}     & \multicolumn{1}{c}{67.3}     & \multicolumn{1}{|c}{62.4}  & \multicolumn{1}{c}{44.6}     & 87.5            \\ 
20     & \multicolumn{1}{c}{47.2}  & \multicolumn{1}{c}{33.7}     & \multicolumn{1}{c}{66.3}     & \multicolumn{1}{|c}{61.2} & \multicolumn{1}{c}{43.6} & 86.2   \\
30    & \multicolumn{1}{c}{\textbf{46.9}}  & \multicolumn{1}{c}{\textbf{33.5}}     & \multicolumn{1}{c}{\textbf{65.9}}     & \multicolumn{1}{|c}{\textbf{60.5}}  & \multicolumn{1}{c}{\textbf{43.1}}     & \textbf{85.6}  \\ \hline
\end{tabular}}
\end{table}
\subsubsection{Number of Denoising Steps}
For faster inference of our diffusion model, we use DDIM \cite{ddim} sampling to accelerate the estimation.
We perform a study on the number of DDIM sampling steps to show its impact.
As presented in Tab. ~\ref{tab:steps}, we assess DiffCap with a subset of denoising steps on the AIST++ and 3DPW-OCC datasets.
Our findings indicate that beyond 5 denoising steps, there is no improvement in pose estimation.
We chose to use 5 DDIM sampling steps as it allowed DiffCap to attain superior performance across most metrics.
\subsubsection{Robustness to Keypoint Noise}
As the 2D keypoints are detected by an off-the-shelf 2D human pose estimator, here we study the robustness of our model against noise in 2D keypoints.
We evaluate the pose results of the previous state-of-the-art fusion-based method RobustCap and our DiffCap on AIST++ dataset by introducing Gaussian noise ($\sigma$) to the 2D keypoints on $Z=1$ plane while preserving the confidence during inference.
\textcolor{red}{As shown in Tab. ~\ref{tab:robustness}, the performance of RobustCap is significantly degraded when introducing more noise to the 2D keypoints, indicating its limited robustness against noise}.
In contrast, DiffCap exhibits less degradation in accuracy, indicating that our method can accurately model human pose even with noisy 2D keypoints input.
\subsubsection{Effectiveness of Diffusion Model}
To evaluate the effectiveness of the Diffusion Model, we replace it with transformer encoders and present the results in Tab. ~\ref{tab:diffusion}.
As shown, Diffusion Model performs better at capturing the human motion, as evidenced by the superior results achieved with our diffusion-based architecture.
\textcolor{red}{We also evaluate the effectiveness of the Pose Diffusion Model by replacing it with transformer encoders that directly estimate human poses from joint positions and inertial signals.
The diffusion model proves beneficial for pose estimation, as it effectively models the human motion prior.}
\subsubsection{Number of Sliding Steps}
We evaluate the effect of the sliding step in the sliding window strategy by setting it to 10, 20, and 30, respectively.
The results are shown in Tab. ~\ref{tab:window-size}.
While reducing the sliding step leads to slightly larger errors, it is much more accurate than previous methods.
\subsection{Live Demo}\label{sec:livedemo}
We use two laptops to run our live demo.
The first laptop runs MediaPipe \cite{mediapipe} to extract the 2D keypoints from each frame of the video stream captured by the camera.
Simultaneously, this laptop receives IMU measurements via Bluetooth and sends both the 2D keypoints and IMU data to the second laptop over a socket connection.
This process is executed on a system with an Intel(R) Core(TM) i7-12700H CPU.
After the second laptop receives the 2D keypoints and IMU data, it runs our model to estimate the human body's joint rotations and global translation (by RobustCap \cite{pan2023fusing}). 
Once the second laptop receives the 2D keypoints and IMU data, it uses our model to estimate the human body's joint rotations and global translation (via RobustCap \cite{pan2023fusing}).
The second laptop then renders the human motion on screen using Unity3D.
This step is carried out on a laptop with an Intel(R) Core(TM) i7-10750H CPU and an NVIDIA RTX 2080 Super graphics card.
For additional hardware, we use Noitom IMUs and a Logitech webcam.
\par
To reduce latency in our live demo, we employ a sliding window of 10 frames, which requires future frames within a duration of 0.32 seconds.
As a result, our live demo runs at 60 fps with a latency of 0.5 seconds (the algorithm latency is 0.32 seconds, while the keypoint detection, network transmission, and rendering add additional 0.18 seconds).
\par 
We showcase our live demo with various motions, including moving in and out of the camera view, severe occlusions (e.g., by clothes, umbrellas, or boards), challenging poses (e.g., dancing), and in-the-wild scenarios.
These results are demonstrated in the accompanying video.
\section{Limitations}
Our DiffCap employs the diffusion model to estimate better poses.
However, we did not apply it to improve translation estimation, which we believe is a promising future work.
Our method exhibits low sensitivity to noise in 2D keypoints as we encode this information into a global condition.
The IMU measurements may be influenced by magnetic disturbances which may happen in the real recording, leading to inaccurate mocap.
Given the visual and inertial signals, how to model the magnetic disturbance and subtract it from the IMU input to avoid the aforementioned problem is still an open question.
\textcolor{red}{Since our approach is purely kinematic-based, the results may exhibit non-physical artifacts such as jitter.
Addressing these limitations through physical optimization or reinforcement learning techniques will be explored in future work.}
\section{Conclusion}
We proposed a novel technique to combine a monocular camera and sparse IMUs to perform human motion capture in real-time.
To the best of our knowledge, this is the first work to use diffusion models to fuse the two modalities of signals in the task of motion capture.
We analyze the characteristics of the two modalities and design different methodologies to handle them.
For the visual signal, considering it may be unavailable on individual frames, we represent it globally through a condition embedding to avoid pollution on local frames.
For the IMU signal, we encode it frame by frame as it is temporally stable and matches the body motion on the temporal domain in a short temporal window.
By the delicate designs, we outperform the existing state-of-the-art techniques on pose accuracy for different scenarios including occlusions and subjects moving out of camera view.
The fusion of the two modalities and learning human motion priors are fluent and seamlessly achieved in our unified diffusion framework. 

\vspace{-33pt}
\begin{IEEEbiography}[{\includegraphics[width=1in,height=1.25in,clip,keepaspectratio]{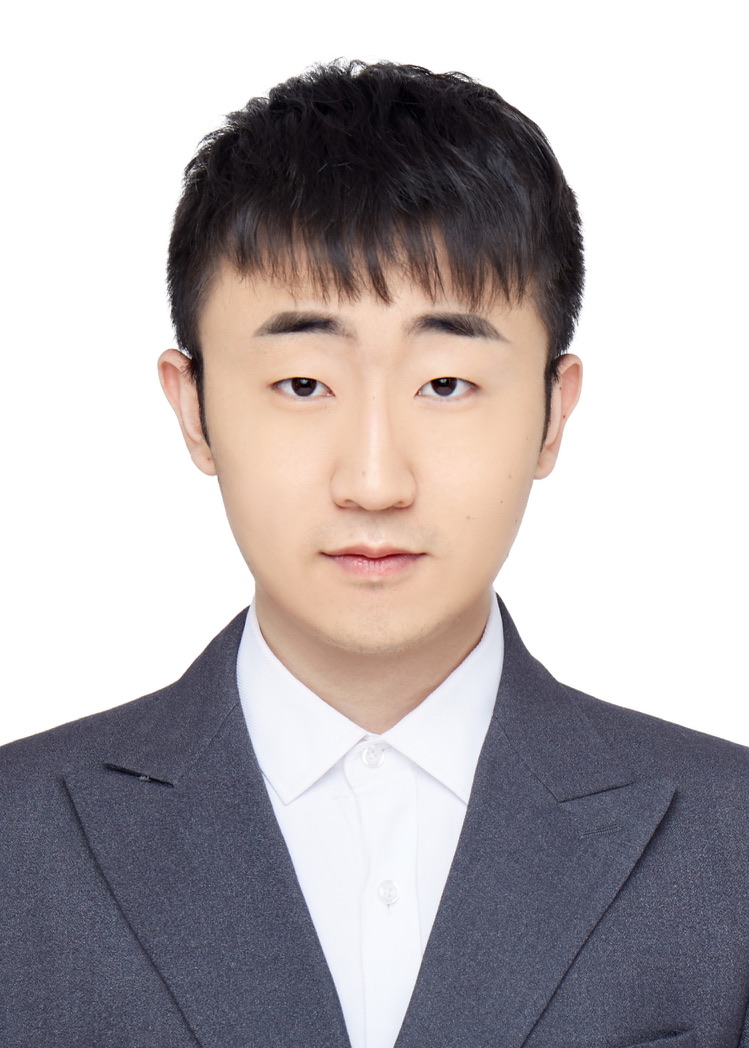}}]{Shaohua Pan} received his B.S. degree in School of Software from Beijing Institute of Technology, Beijing, China, and M.S. degree in School of Software from Tsinghua University, Beijing, China. His research interests include computer vision, computer graphics and generative models.. His research interests include human motion capture, motion generation, and video generation.
\end{IEEEbiography}
\vspace{-33pt}
\begin{IEEEbiography}[{\includegraphics[width=1in,height=1.25in,clip,keepaspectratio]{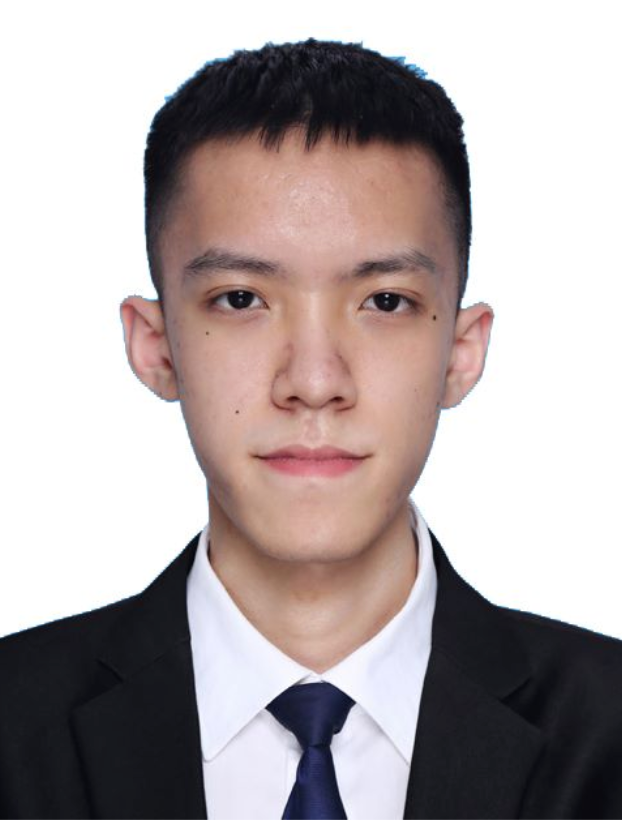}}]{Xinyu Yi} received his B.S. degree in Computer Science from School of the Gifted Young, University of Science and Technology of China, and Ph.D. degree in School of Software from Tsinghua University, Beijing, China. His research interests include human motion capture, and physics-based human-environment interaction.
\end{IEEEbiography}
\vspace{-33pt}
\begin{IEEEbiography}[{\includegraphics[width=1in,height=1.25in,clip,keepaspectratio]{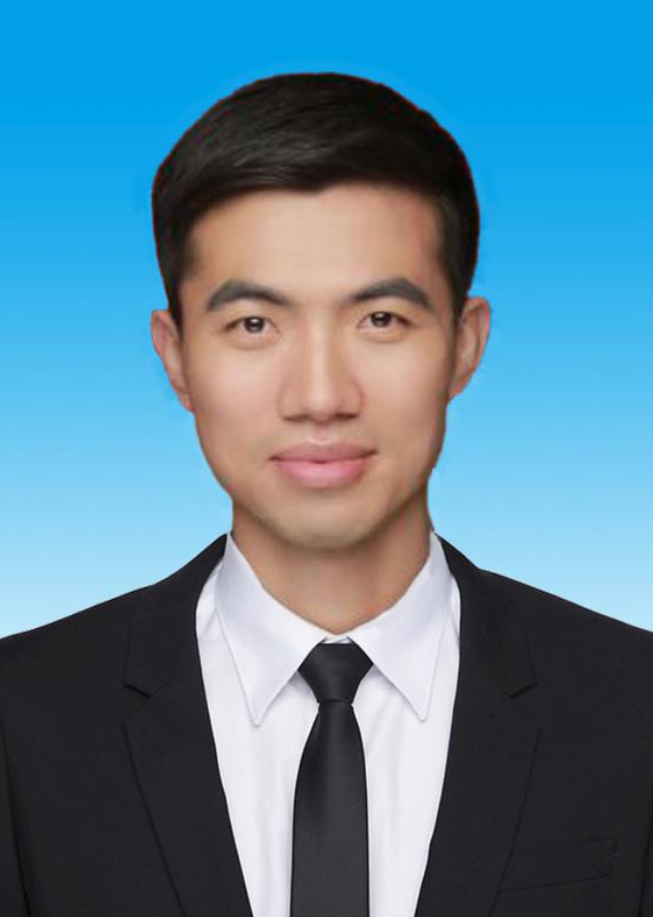}}]{Yan Zhou} received his B.S. and M.S. degrees from the Department of Automation at Tsinghua University in 2013 and 2016, respectively. He is currently a Machine Learning Engineer at Kuaishou Technology (Kwai Inc.). His research interests include human pose estimation, motion generation, and deep generative models.
\end{IEEEbiography}
\vspace{-33pt}
\begin{IEEEbiography}[{\includegraphics[width=1in,height=1.25in,clip,keepaspectratio]{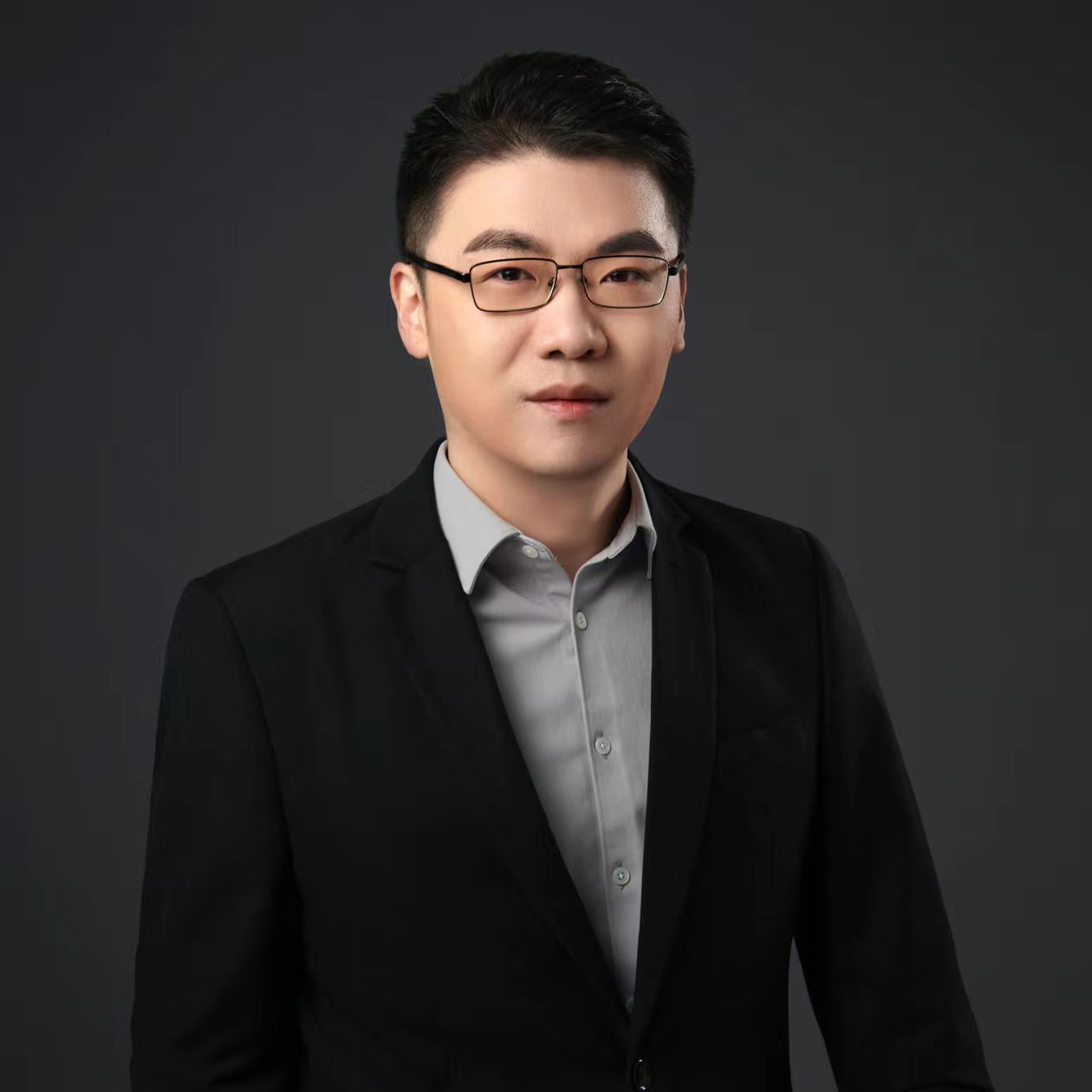}}]{Weihua Jian} is the Manager of the Kuaishou Digital Human Platform, Kuaishou. Weihua has over 10 years of experience in the fields of video, AI, and extended reality. He's currently leading a team at Kuaishou to develop a series of digital human solutions, including the implementation of the Guan Xiaofang IP and the Kuaishou ZhiBo Platform.
\end{IEEEbiography}
\vspace{-33pt}
\begin{IEEEbiography}[{\includegraphics[width=1in,height=1.25in,clip,keepaspectratio]{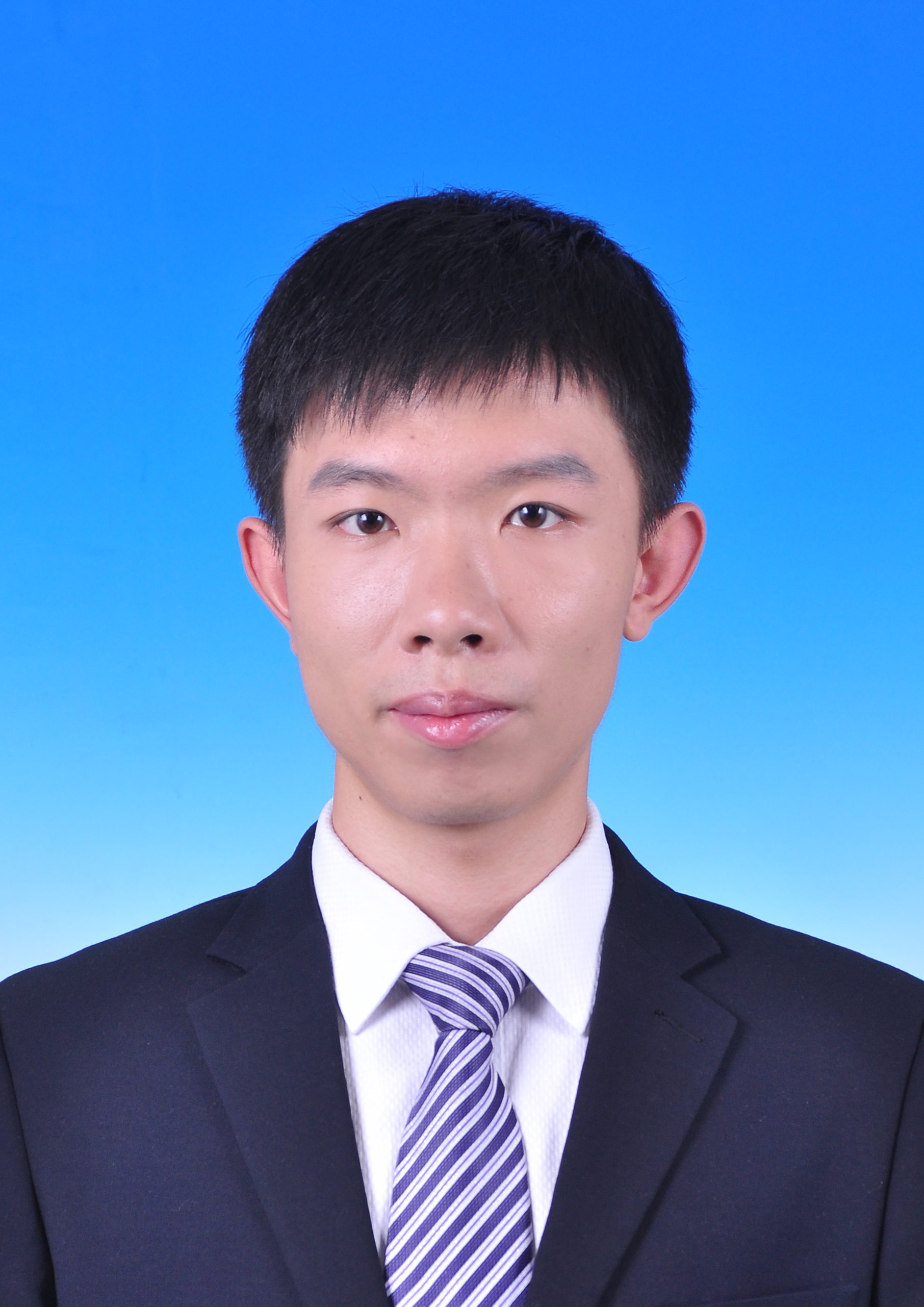}}]{Yuan Zhang} received the B.S. degree in Automation from Beijing University of Posts and Telecommunications in 2009 and the Ph.D. degree in Pattern Recognition from the Institute of Automation, Chinese Academy of Sciences in 2014. His research interests include but are not limited to multimodal understanding and generation, digital human, 3D understanding and reconstruction.
\end{IEEEbiography}
\vspace{-33pt}
\begin{IEEEbiography}[{\includegraphics[width=1in,height=1.25in,clip,keepaspectratio]{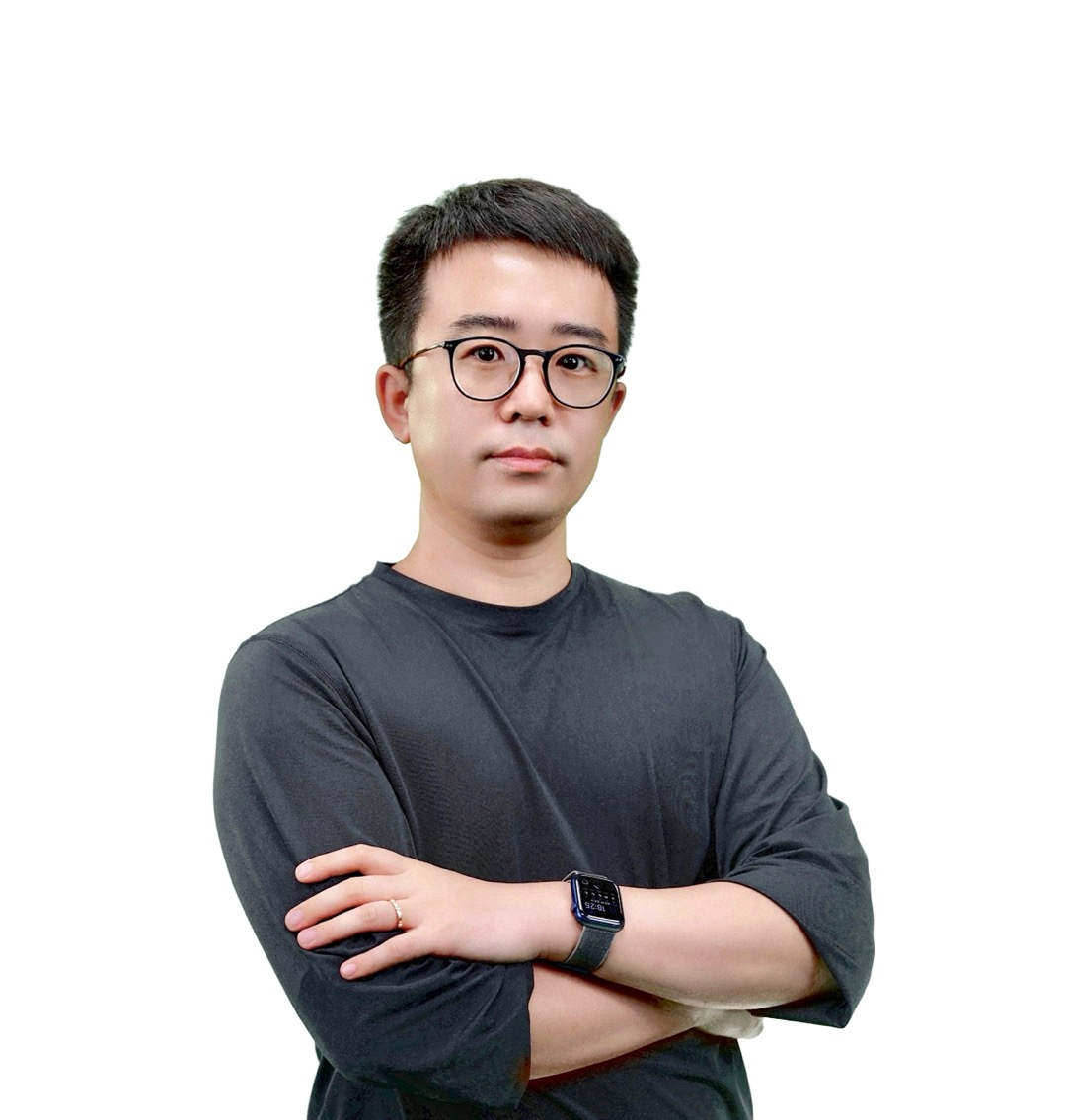}}]{PengFei Wan} received B.E. degree in electronic engineering and information science from the University of Science and Technology of China (USTC), Hefei, China, and Ph.D. degree in electronic and computer engineering from the Hong Kong University of Science and Technology (HKUST), Hong Kong. His research interests include computer vision, computer graphics and generative models.
\end{IEEEbiography}
\vspace{-33pt}
\begin{IEEEbiography}[{\includegraphics[width=1in,height=1.25in,clip,keepaspectratio]{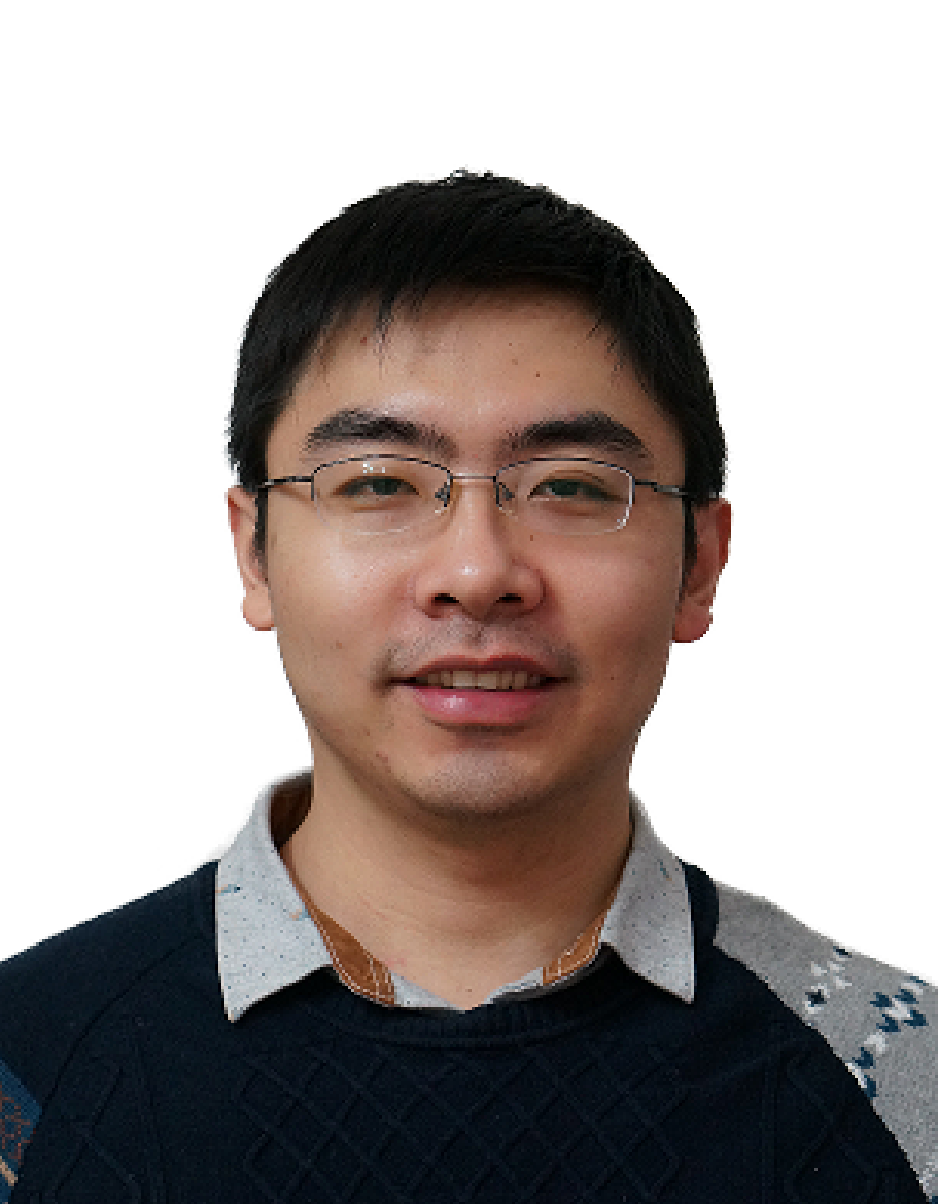}}]{Feng Xu} is currently an associate professor in the School of Software at Tsinghua University. He received B.S. degree in physics from Tsinghua University, Beijing, China in 2007, and Ph.D. degree in automation from Tsinghua University, Beijing, China in 2012. His research interests include facial animation, performance capture, and 3D reconstruction.
\end{IEEEbiography}

\vfill

\end{document}